\documentclass[10pt,twocolumn,letterpaper]{article}

\usepackage{times}
\usepackage{epsfig}
\usepackage{graphicx}
\usepackage{amsmath}
\usepackage{amssymb}

\usepackage[breaklinks=true,bookmarks=false]{hyperref}

\date{\vspace{-5ex}}

\begin{document}

\title{Motion Representation with Acceleration Images}

\author{Hirokatsu Kataoka, Yun He, Soma Shirakabe, Yutaka Satoh\\
National Institute of Advanced Industrial Science and Technology (AIST)\\
Tsukuba, Ibaraki, Japan\\
{\tt\small \{hirokatsu.kataoka, yun.he, shirakabe-s, yu.satou\}@aist.go.jp}
}

\maketitle

\begin{abstract}
Information of time differentiation is extremely important cue for a motion representation. We have applied first-order differential velocity from a positional information, moreover we believe that second-order differential acceleration is also a significant feature in a motion representation. However, an acceleration image based on a typical optical flow includes motion noises. We have not employed the acceleration image because the noises are too strong to catch an effective motion feature in an image sequence. On one hand, the recent convolutional neural networks (CNN) are robust against input noises.

In this paper, we employ acceleration-stream in addition to the spatial- and temporal-stream based on the two-stream CNN. We clearly show the effectiveness of adding the acceleration stream to the two-stream CNN. 
\end{abstract}

\section{Introduction}

Highly discriminative motion representation is needed in the fields of action recognition, event recognition, and video understanding. Space-time interest points (STIP) that capture temporal keypoints are a giant step toward solving visual motion representation. An improvement over STIP is the so-called dense trajectories (DT) proposed by Wang et al.~\cite{WangCVPR2011}.  The simple purpose of DT is to have denser sampling and more various descriptors than STIP. In 2013, DT was improved by three techniques, namely, camera motion estimation with SURF, Fisher vector representation, and detection-based noise canceling~\cite{WangICCV2013}. The powerful framework of DT or improved DT (DT/IDT) has been cited in numerous papers as of 2016. However, the success of convolutional neural networks (CNN) cannot be ignored in image-based recognition. We project motion information into images in order to implement the CNN architecture for motion representation. The two-stream CNN is a noteworthy algorithm to capture the temporal features in an image sequence~\cite{SimonyanNIPS2014}. The integration of spatial and temporal streams allows us to effectively enhance motion representation. We obtain significant knowledge about the spatial information, which helps the temporal feature. The strongest approach introduced is the crosspoint of the IDT and the two-stream CNN. Trajectory-pooled deep-convolutional descriptors (TDD)~\cite{WangCVPR2015} have achieved the highest performance in several benchmarks, such as UCF101~\cite{UCF101} (91.5\%) and HMDB51 (65.9\%)~\cite{HMDB51}. A more recent performance was demonstrated in the ActivityNet challenge in conjunction with CVPR2016. At this performance, the TDD-based approach surprisingly accomplished a 93.2\% mAP  (94.2\% on UCF101 and 69.4\% on HMDB51). 

However, the current approaches heavily rely on the two-stream architecture. To improve motion-based features, we must employ the acceleration stream for richer image representation. In physics, acceleration is the change rate of speed with respect to time. Here, acceleration images are able to extract a precise feature from an image sequence.

\begin{figure*}[t]  
\begin{center}
   \includegraphics[width=1.0\linewidth]{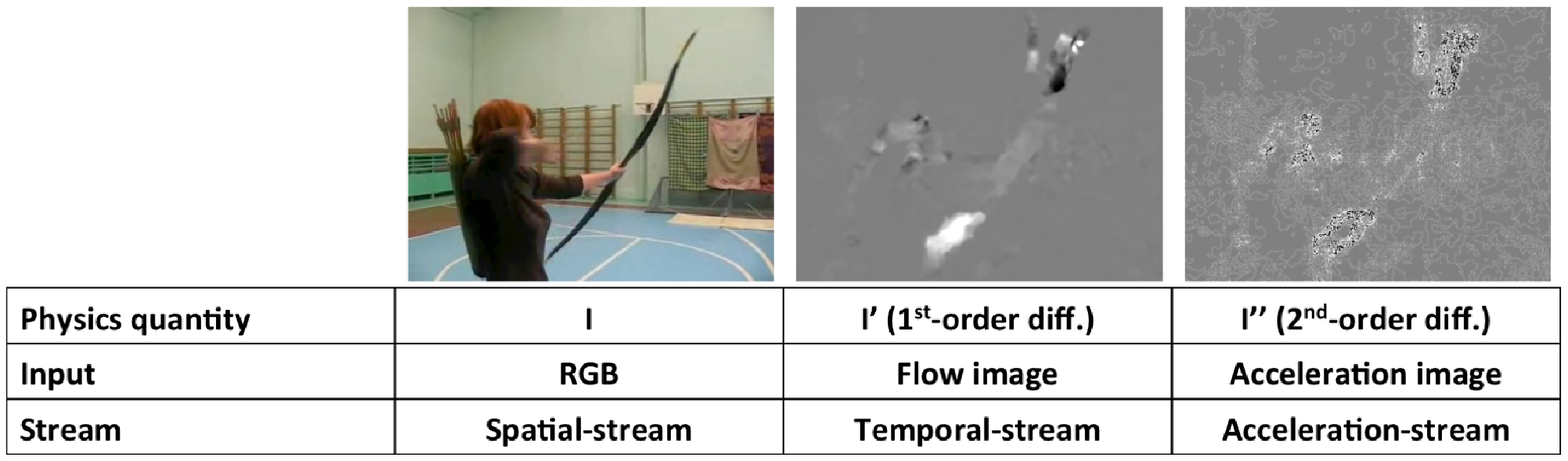}
\end{center}
   \caption{Image representation of RGB ($I$), flow ($I^{'}$), and acceleration ($I^{''}$).}
\label{fig:input}
\end{figure*}

In this paper, we propose the simple technique of using ``acceleration images'' to represent a change of a flow image. The acceleration images must be significant because the representation is different from position (RGB) and speed (flow) images. We apply two-stream CNN~\cite{SimonyanNIPS2014} as the baseline; then, we employ an acceleration stream, in addition to the spatial and the temporal streams. The acceleration images are generated by differential calculations from a sequence of flow images. Although the sparse representation tends to be noisy data (see Figure~\ref{fig:input}), automatic feature learning with CNN can significantly pick up a necessary feature in the acceleration images. We carry out experiments on traffic data in the NTSEL dataset~\cite{KataokaITSC2015}.

\section{Related work}

Space-time interest points (STIP) have been a primary focus in action recognition~\cite{LaptevIJCV2005}. In STIP, time $t$ space is added to the $x,y$ spatial domain. Improvements of STIP have been reported in several papers, such as~\cite{LaptevCVPR2008},~\cite{MarszalekCVPR2009},~\cite{EvertsCVPR2013}. However, the significant approach is arguably the dense trajectories approach (DT)~\cite{WangCVPR2011}. The DT is describes the trajectories that track densely sampled feature points. Descriptors are applied to the densely captured trajectories by histograms of oriented gradients (HOG)~\cite{DalalCVPR2005}, histograms of optical flow (HOF)~\cite{LaptevCVPR2008}, and motion boundary histograms (MBH)~\cite{DalalECCV2006}.

Dense sampling approaches for activity recognition were also proposed in~\cite{JainCVPR2013, KataokaACCV2014, WangICCV2013} after the introduction of the first DT. These studies incremented DT, for example, by eliminating extra flow~\cite{JainCVPR2013} and integrating a higher-order descriptor into the conventional features for fine-grained action recognition~\cite{KataokaACCV2014}. Additionally, Wang \textit{et al.} proposed an IDT~\cite{WangICCV2013} by executing camera motion estimation, canceling detection-based noise, and adding a Fisher vector~\cite{PerronninECCV2010}. More recent work has reported state-of-the-art performance achieved with the concatenation of CNN features and IDT in the THUMOS Challenge~\cite{THUMOS2015, JainTHUMOS2014, ZhuTHUMOS2015}. Jain \textit{et al.} employed a per-frame CNN feature from layers 6, 7, and 8 with AlexNet~\cite{KrizhevskyNIPS2012}. Zhu \textit{et al.}~\cite{ZhuTHUMOS2015} extended both the representations with multi-scale temporal sampling in the IDT~\cite{LanCVPR2015} and video representations in the CNN feature~\cite{XuCVPR2015}. The combination of IDT and CNN synergistically improves recognition performance.

Recently, CNN features with temporal representations have been proposed~\cite{SimonyanNIPS2014, RyooCVPR2015, WangCVPR2015}. Ryoo \textit{et al.} clearly bested IDT+CNN with their pooled time series (PoT) that continuously accumulates frame differences between two frames~\cite{RyooCVPR2015}. The feature is simple but effective for grasping continuous action sequences. The feature type that should be implemented, however, is one that improves the representation so that it adequately fits the transitional action recognition. It is difficult to achieve short-term prediction by using the PoT, because it describes features from a whole image sequence. Kataoka proposed a subtle motion descriptor (SMD) to represent sensitive motion in spatio-temporal human actions~\cite{KataokaBMVC2016}. The SMD enhances the zero-around temporal pooled feature. Two-stream CNN is a well-organized algorithm that captures the temporal feature in an image sequence~\cite{SimonyanNIPS2014}. The integration of the spatial  and the temporal streams allows us to effectively enhance the motion representation. We can obtain significant knowledge about how spatial information helps the temporal feature. Moreover, the strongest approach introduced is at the crosspoint of the IDT and two-stream CNN. TDDs have achieved the highest performance in several benchmarks, such as UCF101 (91.5\%) and HMDB51 (65.9\%)~\cite{WangCVPR2015}.

\section{Acceleration images into two-stream CNN}

\paragraph{Acceleration images.} The placement of acceleration images is shown in Figure~\ref{fig:input}. The acceleration image $I^{''}$ is a second-order differential from a position image $I$ that is RGB input. The acceleration image $I^{''}$ is shown below:
\begin{eqnarray}
   I_{x}^{''} &=& I'(i+1,j) - I'(i,j) \\
   I_{y}^{''} &=& I'(i,j+1) - I'(i,j)
\end{eqnarray}
where $i$ and $j$ are elements of $x$ and $y$. $I'$ indicates a flow image that is calculated with optical flow displacement ($d$)~\cite{SimonyanNIPS2014}:
\begin{eqnarray}
   I_{x}^{'} &=& d^{x}(u,v) \\
   I_{y}^{'} &=& d^{y}(u,v)
\end{eqnarray}
where $(u,v)$ is an arbitrary point.
The acceleration and flow images are stacked 10 frames in a row as $(I_{x1}^{''}, I_{y1}^{''}, I_{x2}^{''}, I_{y2}^{''},...,I_{x10}^{''}, I_{y10}^{''})$ and $(I_{x1}^{'}, I_{y1}^{'}, I_{x2}^{'}, I_{y2}^{'},...,I_{x10}^{'}, I_{y10}^{'})$~\cite{SimonyanNIPS2014}.

We implement VGGNet, which is supported by Limin Wang~\cite{WangarXiv2015}. We integrate the acceleration stream on the two-stream CNN, in addition to the spatial and temporal streams as follows:
\begin{eqnarray}
   f = f_{spa} + \alpha f_{tem} + \beta f_{acc}
\end{eqnarray}
where $f$ indicates the softmax function, and $spa$, $tem$, and $acc$ correspond to the spatial, temporal, and acceleration streams, respectively. $\alpha$ (= 2.0) and $\beta$ (= 2.0) are weighted parameters.

\begin{figure*}[t] 
\begin{center}
   \includegraphics[width=0.95\linewidth]{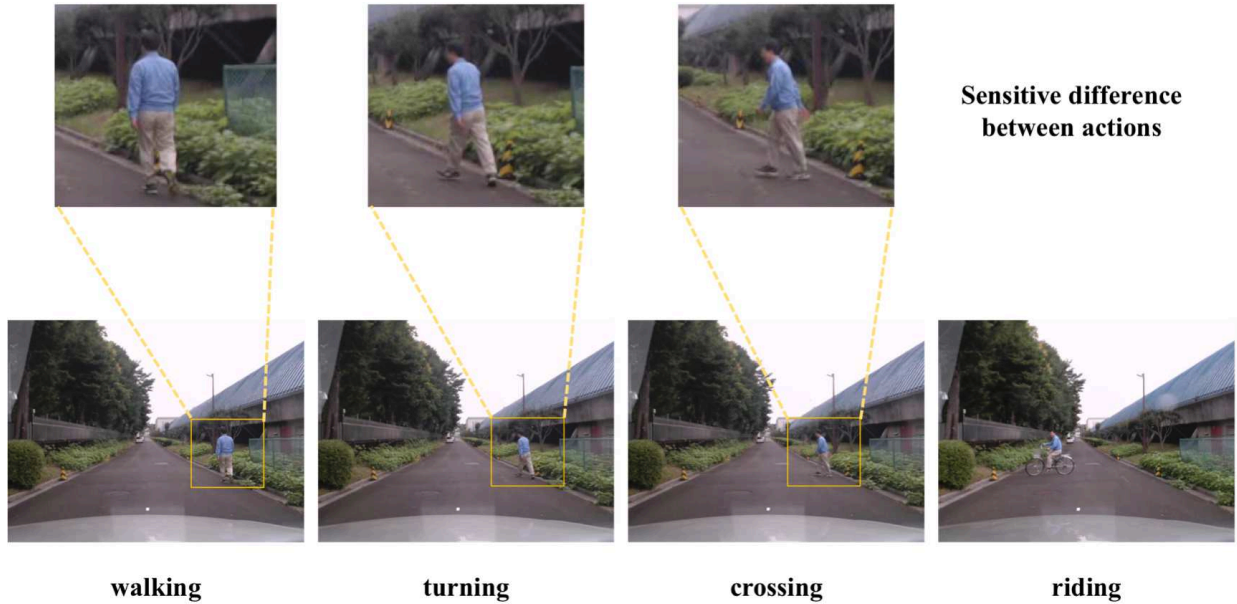}
\end{center}
   \caption{NTSEL dataset.}
\label{fig:dataset}
\end{figure*}

\paragraph{CNN training.} The learning procedure of the spatial and the temporal streams is based on \cite{WangarXiv2015}. We employ a temporal net as a pre-trained model, because the 20-channel input and image values are very similar. The initial learning rate is set as 0.001 and updating is x0.1 for every 10,000 iterations.  The learning of the acceleration stream terminates at 50,000 iterations. We assign a high dropout ratio in all fully connected (fc) layers. We set 0.9 (first fc layer) and 0.9 (second fc layer) for the acceleration stream.

\section{Experiment}

\paragraph{NTSEL dataset (NTSEL)~\cite{KataokaITSC2015} (Figure~\ref{fig:dataset}).} The dataset contains near-miss events captured by a vehicle. We focused on a pedestrian's gradual changes \textit{walking straight, turning}, which is a fine-grained activity on real roads. The four activities are \textit{walking, turning, crossing,} and \textit{riding a bicycle}. The dataset has 100 videos of pedestrian actions. Each of the four actions has 25 videos: 15 videos for training and the other 10 videos for testing. A difficulty of the dataset is to divide walking activities (e.g., \textit{walking, turning, crossing}) with similar appearances from the image sequence. Primitive motion understanding is beneficial to the dataset.

\paragraph{Results.} The results are shown in Table 1. The performance rate is based on per-video calculations. The video recognition system outputs an action label for each video. Our proposed algorithm that adds the acceleration stream significantly outperforms the two-stream CNN with an increase of 2.5\% on the NTSEL dataset. The correct recognitions of the spatial, temporal, and acceleration streams are 87.5\%, 77.5\%, and 82.5\%, respectively. Surprisingly, the acceleration stream performs better than the temporal stream. The acceleration stream effectively recognizes the movement of acceleration in the traffic data. We confirmed that the motion feature of acceleration in an image sequence improves video recognition. Although the knowledge is based on position, speed, and acceleration in physics, we proved the existence of acceleration in the video motion.  Moreover, we believe that the CNN processing automatically selected the dominant feature from the acceleration stream. 

\begin{table}
\begin{center}
\begin{tabular}{lc}
\hline
Approach & \% on NTSEL \\
\hline\hline
Spatial stream & 87.5 \\
Temporal stream & 77.5 \\
Acceleration stream & 82.5 \\
Two streams (S+T)~\cite{WangarXiv2015} & 87.5 \\
Three stream (S+T+A; ours) & \textbf{90.0} \\
\hline
\end{tabular}
\end{center}
\caption{Performance rate of three-stream architecture (spatial + temporal + acceleration; S+T+A) and other approaches on the NTSEL dataset.} 
\end{table}

\section{Conclusion}

In this paper, we propose the definition  of acceleration images that represent a change of a flow image. The acceleration stream is employed as an additional stream to a two-stream CNN. The process of the two-stream CNN picks up a necessary feature in the acceleration images with an automatic feature mechanism. Surprisingly, the motion recognition with the acceleration stream is better than recognition with the temporal stream. 

Our future work is to iteratively differentiate the acceleration images to extract more detailed motions. In particular, we hope to capture more refined features in human motion.

{\small
\bibliographystyle{plain}
\bibliography{acceleration}

\begin{thebibliography}{10}

\bibitem{DalalCVPR2005}
N.~Dalal and B.~Triggs.
\newblock Histograms of oriented gradients for human detection.
\newblock IEEE Conference on Computer Vision and Pattern Recognition (CVPR),
  2005.

\bibitem{DalalECCV2006}
N.~Dalal, B.~Triggs, and C.~Schmid.
\newblock Human detection using oriented histograms of flow and appearance.
\newblock European Conference on Computer Vision (ECCV), 2006.

\bibitem{EvertsCVPR2013}
I.~Everts, J.~C. Gernert, and T.~Gevers.
\newblock Evaluation of color stips for human action recognition.
\newblock IEEE Conference on Computer Vision and Pattern Recognition (CVPR),
  2013.

\bibitem{THUMOS2015}
A.~Gorban, H.~Idrees, Y.-G. Jiang, A.~Roshan~Zamir, I.~Laptev, M.~Shah, and
  R.~Sukthankar.
\newblock {THUMOS} challenge: Action recognition with a large number of
  classes.
\newblock \url{http://www.thumos.info/}, 2015.

\bibitem{JainTHUMOS2014}
M.~Jain, J.~Gemert, and C.~G.~M. Snoek.
\newblock University of amsterdam at thumos challenge2014.
\newblock European Conference on Computer Vision Workshop (ECCVW), 2014.

\bibitem{JainCVPR2013}
M.~Jain, H.~Jegou, and P.~Bouthemy.
\newblock Better exploiting motion for better action recognition.
\newblock IEEE Conference on Computer Vision and Pattern Recognition (CVPR),
  2013.

\bibitem{KataokaITSC2015}
H.~Kataoka, Y.~Aoki, Y.~Satoh, S~Oikawa, and Y.~Matsui.
\newblock Fine-grained walking activity recognition via driving recorder
  dataset.
\newblock IEEE Intelligent Transportation Systems Conference (ITSC), 2015.

\bibitem{KataokaACCV2014}
H.~Kataoka, K.~Hashimoto, K.~Iwata, Y.~Satoh, N.~Navab, S.~Ilic, and Y.~Aoki.
\newblock Extended co-occurrence hog with dense trajectories for fine-grained
  activity recognition.
\newblock Asian Conference on Computer Vision (ACCV), 2014.

\bibitem{KataokaBMVC2016}
H.~Kataoka, Y.~Miyashita, M.~Hayashi, K.~Iwata, and Y.~Satoh.
\newblock Recognition of transitional action for short-term action prediction
  using discriminative temporal cnn feature.
\newblock British Machine Vision Conference (BMVC), 2016.

\bibitem{KrizhevskyNIPS2012}
A.~Krizhevsky, I.~Sutskever, and G.~E. Hinton.
\newblock Imagenet classification with deep convolutional neural networks.
\newblock Neural Information Processing Systems (NIPS), 2012.

\bibitem{HMDB51}
H.~Kuehne, H.~Jhuang, E.~Garrote, T.~Poggio, and T.~Serre.
\newblock {HMDB}: a large video database for human motion recognition.
\newblock International Conference on Computer Vision (ICCV), 2011.

\bibitem{LanCVPR2015}
Z.~Lan, M.~Lin, X.~Li, A.~G. Hauptmann, and B.~Raj.
\newblock Beyond gaussian pyramid: Multi-skip feature stacking for action
  recognition.
\newblock IEEE Conference on Computer Vision and Pattern Recognition (CVPR),
  2015.

\bibitem{LaptevIJCV2005}
I.~Laptev.
\newblock On space-time interest points.
\newblock International Journal of Computer Vision (IJCV), 2005.

\bibitem{LaptevCVPR2008}
I.~Laptev, M.~Marszalek, C.~Schmid, and B.~Rozenfeld.
\newblock Learning realistic human actions from movies.
\newblock IEEE Conference on Computer Vision and Pattern Recognition (CVPR),
  2008.

\bibitem{MarszalekCVPR2009}
M.~Marszalek, I.~Laptev, and C.~Schmid.
\newblock Actions in context.
\newblock IEEE Conference on Computer Vision and Pattern Recognition (CVPR),
  2009.

\bibitem{PerronninECCV2010}
F.~Perronnin, J.~Sanchez, and T.~Mensink.
\newblock Improving the fisher kernel for large-scale image classification.
\newblock European Conference on Computer Vision (ECCV), 2010.

\bibitem{RyooCVPR2015}
M.~S. Ryoo, B.~Rothrock, and L.~Matthies.
\newblock Pooled motion features for first-person videos.
\newblock IEEE Conference on Computer Vision and Pattern Recognition (CVPR),
  2015.

\bibitem{SimonyanNIPS2014}
K.~Simonyan and A.~Zisserman.
\newblock Two-stream convolutional networks for action recognition.
\newblock Neural Information Processing Systems (NIPS), 2014.

\bibitem{UCF101}
K.~Soomro, A.~R. Zamir, and M.~Shah.
\newblock Ucf101: A dataset of 101 human action classes from videos in the
  wild.
\newblock CRCV-TR-12-01, 2012.

\bibitem{WangCVPR2011}
H.~Wang, A.~Klaser, C.~Schmid, and L.~Cheng-Lin.
\newblock Action recognition by dense trajectories.
\newblock IEEE Conference on Computer Vision and Pattern Recognition (CVPR),
  2011.

\bibitem{WangICCV2013}
H.~Wang and C.~Schmid.
\newblock Action recognition with improved trajectories.
\newblock International Conference on Computer Vision (ICCV), 2013.

\bibitem{WangCVPR2015}
L.~Wang, Y.~Qiao, and X.~Tang.
\newblock Action recognition with trajectory-pooled deep-convolutional
  descriptors.
\newblock IEEE Conference on Computer Vision and Pattern Recognition (CVPR),
  2015.

\bibitem{WangarXiv2015}
L.~Wang, Y.~Xiong, Z.~Wang, and Y.~Qiao.
\newblock Towards good practices for very deep two-stream convnets.
\newblock arXiv pre-print 1507.02159, 2015.

\bibitem{XuCVPR2015}
Z.~Xu, Y.~Yang, and A.~G. Hauptmann.
\newblock A discriminative cnn video representation for event detection.
\newblock IEEE Conference on Computer Vision and Pattern Recognition (CVPR),
  2015.

\bibitem{ZhuTHUMOS2015}
L.~Zhu, Y.~Yang, and A.~G. Hauptmann.
\newblock Uts-cmu at thumos 2015.
\newblock CVPR2015 International Workshop and Competition on Action Recognition
  with a Large Number of Classes, 2015.

\end{thebibliography}
}

\end{document}